\documentclass{INTERSPEECH2023}

\interspeechcameraready

\title{An Information-Theoretic Analysis \\ of Self-supervised Discrete Representations of Speech}

\name{Badr M. Abdullah, Mohammed Maqsood Shaik, Bernd M\"obius, Dietrich Klakow}
\address{
Language Science and Technology (LST), Saarland University, Germany \\
Saarland Informatics Campus, Germany
  }
\email{\{babdullah|mmshaik|moebius|dietrich\}@lsv.uni-saarland.de}

\begin{document}

\maketitle
 
\begin{abstract}
Self-supervised representation learning for speech often involves a quantization step that transforms the acoustic input into discrete units. 
However, it remains unclear how to characterize the relationship between these discrete units and abstract phonetic categories such as phonemes. 
In this paper, we develop an information-theoretic framework whereby we represent each phonetic category as a distribution over discrete units. 
We then apply our framework to two different self-supervised models (namely, wav2vec 2.0 and XLSR) and use American English speech as a case study. 
Our study demonstrates that the entropy of phonetic distributions reflects the variability of the underlying speech sounds, with phonetically similar sounds exhibiting similar distributions. 
While our study confirms the lack of direct one-to-one correspondence, we find an intriguing indirect relationship between phonetic categories and discrete units.

\end{abstract}
\noindent\textbf{Index Terms}: discrete speech representations, self-supervised learning, information theory

\section{Introduction}

Self-supervised learning (SSL) for the speech modality is an active area of research that aims to develop models that build meaningful speech representations from raw audio without any explicit labels or transcriptions (see \cite{9893562} for an overview). 
These models can be further adapted for downstream tasks such as automatic speech recognition and speaker identification, and have become the state-of-the-art approach even when limited labeled data are available \cite{van2018representation, Schneider2019wav2vecUP, baevski2020wav2vec, hsu2021hubert}. %
Recently, it has become a common practice to include a quantization module within the architecture of SSL speech models that transforms the acoustic input into a sequence of discrete entities. 
Besides representing the complex acoustic signal in a compact and computationally efficient manner, learning discrete representations of speech can also facilitate training large SSL speech models using a masked language modeling objective similar to those employed in natural language processing (e.g., BERT \cite{devlin-etal-2019-bert}). 

Nevertheless, the nature of the discrete units learned via self-supervision remains an under-explored area of research. 
A key question is whether these discrete representations correspond to abstract phonetic categories such as phonemes.
A few recent studies have investigated the discrete units from a neural network interpretability point of view \cite{higy2021discrete, nguyen2022discrete, wells2022phonetic, sicherman2023analysing}.
The analysis in \cite{wells2022phonetic} showed that the discrete units correspond to low-level ``sub-phonetic'' events---rather than high-level phonetic categories---since they are sensitive to context-dependent and non-phonemic variations in speech. 
In \cite{sicherman2023analysing}, the authors concluded that there exists a strong correspondence between discrete units and phonemes, and attributed the lack of consistent phoneme-to-unit mapping to variations in phonological contexts. 
These findings seem to be contradictory and rely on different definitions of the term ``phoneme'', and thus remain inconclusive.

\begin{figure}[t]
  \centering
  \includegraphics[width=0.99\linewidth]{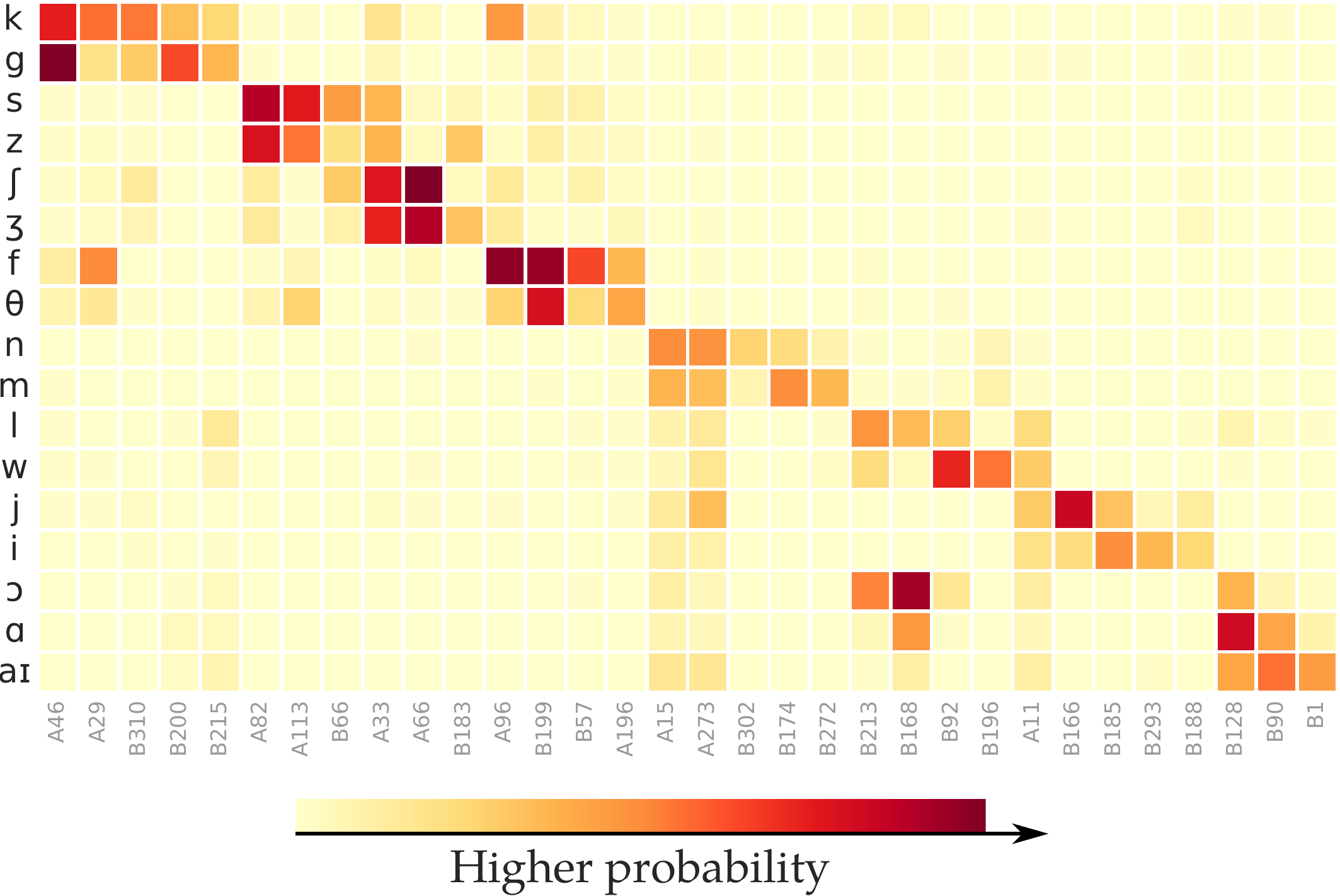}
  \caption{Phonetic categories as empirical probability  distributions over discrete units of the multilingual \textsc{xlsr} model. }
  \label{fig:fig_1}
\end{figure}

Although information theory was initially proposed as a mathematical theory of communication \cite{shannon2001mathematical}, it also provides a quantitative framework for measuring the amount of information conveyed by linguistic units, such as words or sounds. 
Information theory has been adopted as a framework to study various aspects of linguistic structure, including phonology \cite{10.1162/tacl_a_00296, pimentel-etal-2021-surprisal}, morphology \cite{rathi-etal-2021-information, wu-etal-2019-morphological}, and syntax \cite{Hahn2018AnIE, futrell-etal-2015-quantifying}. 
In this paper, we build on this line of research and develop information-theoretic metrics to analyze the correspondence between phonetic categories  and discrete units. 
Concretely, we make the following contributions: 

\begin{itemize}[] %
    \item We develop an empirical approach to represent each phonetic category as a probability distribution over discrete units using two self-supervised pre-trained models: English wav2vec 2.0 (henceforth \textsc{w2v2}) and multilingual wav2vec-XLSR (henceforth \textsc{xlsr}) (\S2).
    \item  We characterize each phonetic category using the notion of information entropy and demonstrate that entropy quantifies acoustic-phonetic variability (\S4).
    \item  We quantify the dissimilarity between phonetic distributions using Jensen-Shannon divergence and illustrate that this metric highly reflects feature-based phonetic similarity (\S5). 
\end{itemize}

\section{Research methodology}

\subsection{Speech quantization via self-supervised learning}
Consider a continuous acoustic signal represented as a sequence of $T$ acoustic frames  $\mathbf{x} = (\mathbf{x}_1, \dots, \mathbf{x}_T)$. 
Here, $\mathbf{x}_t$ could either be an interval of the raw waveform or a spectral vector such as MFCCs. 
Given a pre-trained speech encoder, the signal $\mathbf{x}$ is first transformed via a local, temporal convolutional encoder $\mathcal{F}: \mathcal{X} \mapsto \mathcal{Z}$ into a sequence of latent speech representations in a continuous space as $\mathcal{F}(\mathbf{x}) = \mathbf{z} = (\mathbf{z}_1, \dots, \mathbf{z}_T)$, where $\mathbf{z}_t \in \mathbb{R}^{d} $. 
As a part of the quantization step, the sequence of continuous representations gets discretized to produce a sequence of discrete units $\mathcal{D}(\mathbf{z}) = \boldsymbol{\omega} = (\omega_1, \dots, \omega_T)$, where  $\mathcal{D}: \mathcal{Z} \mapsto \Omega$ is a vector-to-centroid mapping and $\omega_t \in \Omega$ is the index of the centroid. 
Here, we use $\Omega$ to denote the finite set of discrete units within the model codebook.
During pre-training using masked learning objectives, the corresponding quantized representations of these discrete units become the targets of the model prediction. 

\subsection{Phonetic categories as distributions over discrete units}
Consider a speech corpus that is transcribed and aligned to phonetic segments given an inventory of phonetic categories $\Phi$. 
In this scenario, a phonetic category can be considered as a set of $K$ different acoustic exemplars obtained from the corpus,  $\boldsymbol{\varphi} = \{\varphi^{1},\dots, \varphi^{K}\}$. 
These exemplars represent different acoustic realizations of the underlying phonetic category, and should optimally be produced by various speakers in diverse phonological contexts. 
Using the feature encoder and quantization module of a self-supervised speech model, we transform the associated acoustic segments of all exemplars $\{\mathbf{x}^{1},\dots, \mathbf{x}^{K}\}$ into a discrete representation to obtain a collection of discrete sequences $\{(\omega^{1}_1, \dots, \omega^{1}_{\tau_1}), \dots, (\omega^{k}_1, \dots, \omega^{k}_{\tau_k})\}$ for each phonetic category. 
We then discard the exemplar identity as well as the sequential nature of each discrete sequence and view each phonetic category as a bag of discrete units.  
In this approach, each phonetic category can be described as a frequency distribution over the units in $\Omega$.
To facilitate our information-theoretic analysis, we turn the frequency distribution into a probability  distribution  where the probability of observing a discrete unit $\omega$ under a phonetic category  $\boldsymbol{\varphi}$ is calculated using maximum likelihood estimation  as follows 
\begin{equation}
    \boldsymbol{p}_\varphi(\omega_i) = \frac{N_\varphi(\omega_i)}{\sum_{\pi \in \Omega}{N_\varphi(\pi)}}
\end{equation}
Here, $N_\varphi: \Omega \mapsto \mathbb{Z}^{+} $ is a function that returns the number of occurrences of a discrete unit under the phonetic category $\boldsymbol{\varphi}$, and therefore
$\boldsymbol{p}_\varphi: \Omega \mapsto [0, 1] $ is a probability mass function defined over $\Omega$ such that $\sum_{\omega \in \Omega}{\boldsymbol{p}_\varphi(\omega)} = 1$. 
Note that each phonetic category in our analysis has its own $\boldsymbol{p}_\varphi$ and $N_\varphi$ functions. 
For example, the vowels /\ae/ and /\textopeno / are represented as two empirical distributions $\boldsymbol{p}_\text{/\ae/}$ and $\boldsymbol{p}_\text{/\textopeno/}$, respectively. 
Given our representation of a phonetic category as a distribution over discrete units $\boldsymbol{p}_\varphi$, we can employ information-theoretic metrics to characterize each phonetic distribution. 
For simplicity, we henceforth omit the subscript notation in $\boldsymbol{p}_\varphi$ and use $\boldsymbol{p}$ to denote a  distribution associated with a single phonetic category.

\section{Experimental data and models}
\textbf{Speech data.} \hspace{0.25cm} We use the TIMIT speech corpus which consists of recordings from 630  American English speakers each speaking 10 different sentences, for a total of 6,300 sentences covering a diverse range of ages, genders, and regional accents from across the United States \cite{garofolo1993timit}.
Following \cite{rasanen2016analyzing}, the original  phonetic categories of TIMIT annotation are mapped to the reduced set of 40 categories.
We exclude silences  and closures from our analysis. 
\vspace{0.2cm}

\noindent
\textbf{SSL speech models.} \hspace{0.25cm} We conduct our analysis using two publicly available (via the HuggingFace Model Hub) SSL speech models: (1) monolingual English wav2vec 2.0-\textsc{base} \cite{baevski2020wav2vec}, which is a 12-layer transformer model, and (2) multilingual wav2vec XLSR-53-\textsc{large} \cite{conneau2020unsupervised}, which is a 24-layer transformer model trained on different languages. 
Both models employ two codebooks with 320 discrete units each, for a total of 640 units in each model. 
We consider the concatenation of the two codebooks as the set of discrete units in our analysis, thus $|\Omega| = 640$.
\vspace{0.2cm}

\noindent
\textbf{Code and reproducibility.} \hspace{0.25cm} Our analysis code is publicly available on GitHub\footnote{\url{https://github.com/uds-lsv/phone2unit}}.

\section{Analysis I: Phonetic variability as information entropy}
\label{sec_entropy}
\subsection{Information content and entropy} 
For any discrete unit within the codebook $\omega \in \Omega$, we measure its information content, or surprisal under a specific phonetic category as
\begin{equation}
     \eta({\omega}) = - \text{log}_2 \, \boldsymbol{p}(\omega)
\end{equation}
which quantifies the unexpectedness of the discrete unit to be observed under the phonetic category associated with the distribution $\boldsymbol{p}$. It is measured in bits.
The uncertainty or ``randomness'' of the distribution $\boldsymbol{p}$ can be quantified as the average surprisal, or entropy 
\begin{equation}
    \label{eq_entropy}
     H(\boldsymbol{p}) = \sum_{\omega \in \Omega} \boldsymbol{p}(\omega) \, \eta({\omega})
\end{equation}
where $0 \leq H(\boldsymbol{p}) \leq \text{log}_2|\Omega|$.
If all acoustic realizations of a  phonetic category are associated with a single discrete unit, then its entropy is minimal $H(\boldsymbol{p}) = 0$. 
On the other hand, a distribution of a phonetic category is maximally entropic (i.e., $H(\boldsymbol{p}) = \text{log}_2|\Omega|$) when all discrete units are equally likely to be aligned to this category. 
Therefore, entropy can be viewed as a measure of (within-category) acoustic-phonetic variability in our case. 
That is, the more entropic a phonetic category is, the higher the difficulty of predicting its alignment to discrete units. 
Note that our measure of variability is similar to the  measure of diversity (i.e., the unit purity measure) introduced in \cite{hsu2021hubert}, but we express the variability of phonetic distributions using information-theoretic metrics.

\begin{figure}[t]
  \centering
  \includegraphics[width=0.90\linewidth]{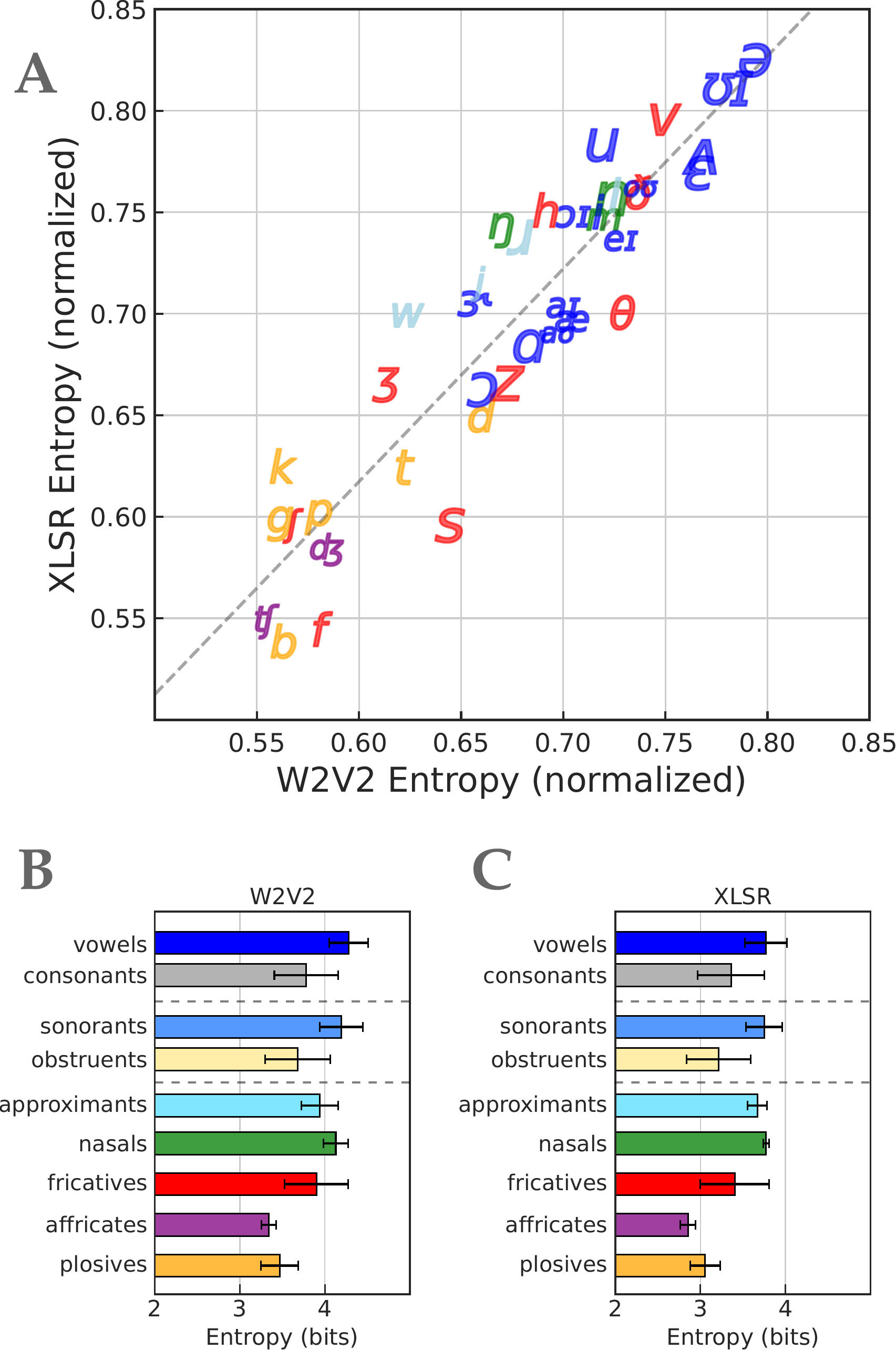}
  \caption{(A) The (normalized) entropy of  each phonetic category in \textsc{w2v2} ($x$-axis) vs \textsc{xlsr} ($y$-axis). (B-C) The entropy of several selected articulatory classes in \textsc{w2v2} (B) and \textsc{xlsr} (C).}
  \label{fig:all_entropy}
\end{figure}

\noindent
\subsection{Entropy per phonetic category}
We compute the entropy of each phonetic category using Eq.~\ref{eq_entropy}.
First, we find that phonetic categories are more entropic on average under \textsc{w2v2} (mean $ H=3.97$) compared to \textsc{xlsr} (mean $ H=3.52$). 
After inspecting the phone-to-unit alignment of the TIMIT corpus, we attribute this behavior to different utilization of the codebooks across the two models.
While there are $56.6\%$ of the discrete units under \textsc{w2v2} with non-zero counts across all phonetic categories, only $24.2\%$ of the units have non-zero counts under \textsc{xlsr}. 
This difference gets reflected in lower entropy values in \textsc{xlsr} compared to \textsc{w2v2}.

Fig.~\ref{fig:all_entropy} illustrates the results of our analysis with entropy as a measure of phonetic variability.  
Fig.~\ref{fig:all_entropy}A shows the entropy of each phonetic category in \textsc{w2v2} ($x$-axis) and \textsc{xlsr} ($y$-axis). 
We report the normalized entropy in Fig.~\ref{fig:all_entropy}A to account for differences in entropy values between the two models. 
 In addition, we group phonetic categories according to several articulatory classes, average the entropy over the categories within each class, and depict the result for \textsc{w2v2} (Fig.~\ref{fig:all_entropy}B) and \textsc{xlsr} (Fig.~\ref{fig:all_entropy}C). 
From Fig.~\ref{fig:all_entropy}A, we observe a strong correlation between the two models (Pearson's $r = 0.92, p \ll 0.001$).
When considering entropy values, we see that none of the phonetic categories is minimally entropic (i.e, $H(\boldsymbol{p}) = 0$), which confirms the findings in the literature about the lack of one-to-one correspondence between high-level abstract phonetic categories and discrete units in self-supervised speech models.

Regarding the variation of entropy across different phonetic categories, we observe that vowels tend to be more entropic than consonants in \textsc{w2v2} ($H_V = 4.28 > H_C = 3.78$) and \textsc{xlsr} ($H_V = 3.77 > H_C = 3.36$). 
This reflects a higher variability in the acoustic realizations of vowels compared to consonants, since vowels are subject to a higher degree of variation due to vowel reduction in unstressed syllables and co-articulation, as well as other factors such as cross-speaker and dialect variability \cite{peterson1952control, hillenbrand1995acoustic, hagiwara1997dialect}. 
For consonants, the nasal sounds (i.e., \textipa{/n, m, \ng/}) are the most entropic consonant group, followed by the approximant sounds (i.e., \textipa{/l, j, w, \textturnr/}), and then by the fricative sounds (i.e., \textipa{/D, z, Z, v, T, s, S, f, h/}).  
We also observe that resonating consonants (i.e., nasals and approximants) exhibit higher variability on average than obstruents (i.e., plosives, fricatives, and affricates). 
Furthermore, we find an effect of voicing on variability since the voiced fricatives (i.e., \textipa{/D, z, Z, v/}) are more entropic than their voiceless counterparts (i.e., \textipa{/T, s, S, f/}). 
For example, consider the voiceless-voiced contrast \textipa{/f-v/} where \textipa{/v/} is substantially more entropic than \textipa{/f/} under \textsc{w2v2} ($H(\text{\textipa{/v/}}) = 4.40 > H(\text{\textipa{/f/}}) = 3.41$) and \textsc{xlsr} ($H(\text{\textipa{/v/}}) = 4.01 > H(\text{\textipa{/f/}}) = 2.75$).
This effect of voicing can be explained by the presence of low-frequency voicing energy in voiced fricatives which is likely to vary due to cross-speaker variability. 
Finally, the affricates (i.e., \textipa{/dZ, tS/}) are found to be the least entropic consonant category under both \textsc{w2v2} ($H = 3.33$) and \textsc{xlsr} ($H = 2.86$).

\section{Analysis II: Phonetic dissimilarity as Jensen-Shannon divergence}

\begin{figure*}[t]
  \centering
  \includegraphics[width=0.8\linewidth]{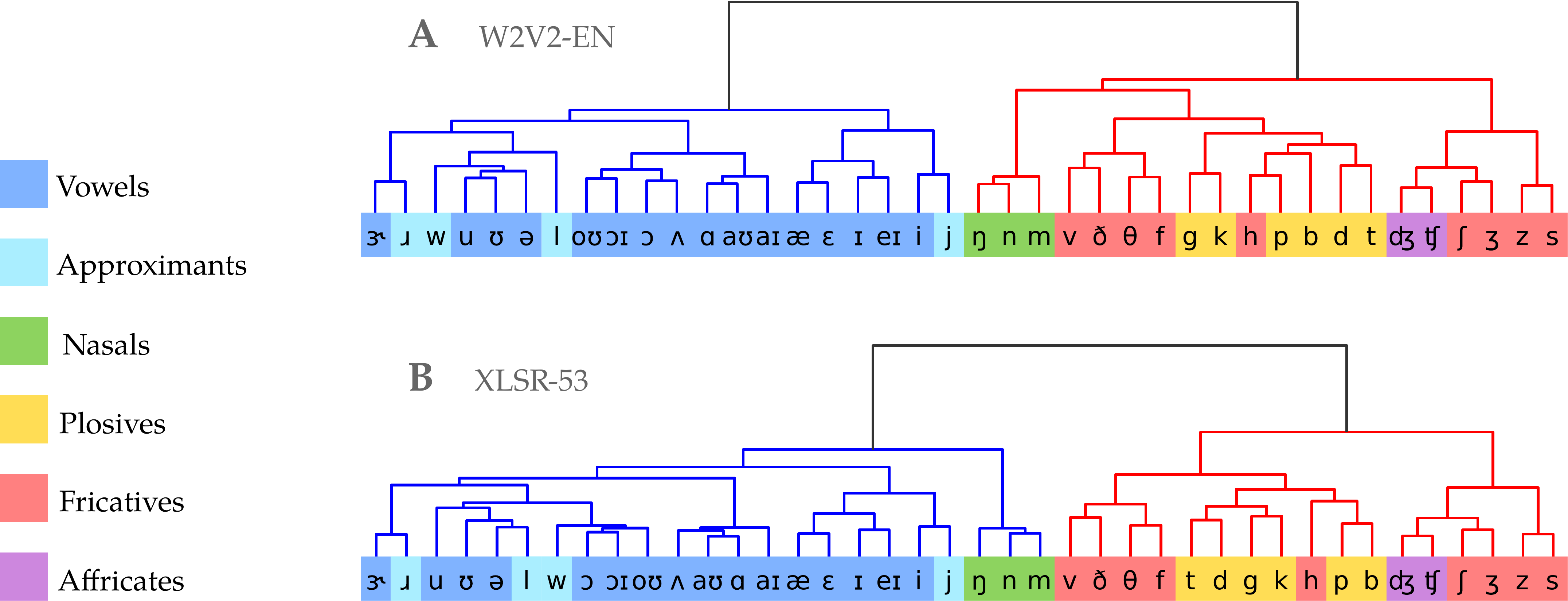}  
  \caption{The resulting clusters from applying agglomerative hierarchical clustering over the distance matrix, where our measure of the distance is the Jensen-Shannon divergence between phonetic distributions: (A) \textsc{w2v2}  and (B) \textsc{xlsr}.}
  \label{fig:clustering}
\end{figure*}

\subsection{Relative entropy and divergence} 
Consider two phonetic distributions $\boldsymbol{p}$ and $\boldsymbol{q}$ that are defined over the same set of discrete units $\Omega$. 
To quantify how different $\boldsymbol{p}$ is from $\boldsymbol{q}$, we measure the expected surprisal from using $\boldsymbol{q}$ as a model distribution  when the true distribution is  $\boldsymbol{p}$.
This quantity is known as the relative entropy or Kullback–Leibler divergence
\begin{equation}
     D_{KL}(\boldsymbol{p} \, || \, \boldsymbol{q}) = - \sum_{\omega \in \Omega} \boldsymbol{p}(\omega) \, \text{log}_2 \, \frac{\boldsymbol{q}(\omega)}{\boldsymbol{p}(\omega)}
\end{equation}
Here, $D_{KL}(\boldsymbol{p} \, || \, \boldsymbol{q}) \geq 0$, with $D_{KL}(\boldsymbol{p} \, || \, \boldsymbol{q}) = 0$ only if $\boldsymbol{p} = \boldsymbol{q}$. 
Note that relative entropy is not symmetric, that is, $D_{KL}(\boldsymbol{p} \, || \, \boldsymbol{q}) \neq D_{KL}(\boldsymbol{q} \, || \, \boldsymbol{p})$.
Since a symmetric metric is more suitable for our analysis, we therefore measure the distance between two probability distributions using Jensen-Shannon divergence (JSD)
\begin{equation}
     D_{JS}(\boldsymbol{p} \, || \, \boldsymbol{q}) = \frac{1}{2}D_{KL}(\boldsymbol{p} \, || \, \boldsymbol{m}) + \frac{1}{2}D_{KL}(\boldsymbol{q} \, || \, \boldsymbol{m})
\end{equation}
where $\boldsymbol{m} = \frac{1}{2} \, \boldsymbol{p} + \boldsymbol{q}$ and $0 \leq D_{JS}(\boldsymbol{p} \, || \, \boldsymbol{q})  \leq 1$. 
Here, our goal is to investigate the degree to which the distance between distributions reflects phonetic similarity. 
Therefore, we use JSD as a measure of phonetic (dis)similarity in our analysis.

\subsection{Exploratory similarity analysis}

\begin{table}[b]
\centering
\caption{Top-5 most similar phonetic categories to each of the categories  \textipa{/w, E, S, g/} in both \textsc{w2v2} (\textsc{w}) and \textsc{xlsr} (\textsc{x}). }
\label{tab:similar_phones}
\begin{tabular}{@{}ccccccccc@{}}
\toprule
 & \multicolumn{2}{c}{\textipa{/w/}} & \multicolumn{2}{c}{\textipa{/E/}} & \multicolumn{2}{c}{\textipa{/S/}} & \multicolumn{2}{c}{\textipa{/g/}} \\ \midrule %
  & \textsc{w}  & \textsc{x} & \textsc{w}  & \textsc{x} & \textsc{w}  & \textsc{x} & \textsc{w} & \textsc{x}   \\ \midrule
1 & \textipa{/l/}  & \textipa{/l/} &  \textipa{/\ae/} & \textipa{/\ae/}  & \textipa{/tS/}  & \textipa{/tS/}  & \textipa{/k/}  &  \textipa{/k/}    \\
2 &  /u/  & \textipa{/u/} & \textipa{/2/}  &  \textipa{/I/} & \textipa{/Z/}  & \textipa{/Z/}  &  \textipa{/b/} & \textipa{/b/}     \\
3 & \textipa{/U/} & \textipa{/U/} &  \textipa{/I/} &  \textipa{/2/} & \textipa{/dZ/}  & \textipa{/dZ/}  & \textipa{/d/}  &  \textipa{/d/}   \\
4 & \textipa{/O/} & \textipa{/@/} & \textipa{/eI/}  &  \textipa{/eI/} &  \textipa{/s/} & \textipa{/s/}  & \textipa{/p/}  & \textipa{/p/}   \\
5 & \textipa{/OI/} & \textipa{/oU/} &  \textipa{/aU/} & \textipa{/aI/}  & \textipa{/z/}  & \textipa{/z/}  & \textipa{/D/} & \textipa{/h/}   \\ \bottomrule
\end{tabular}
\end{table}

Table \ref{tab:similar_phones} presents a qualitative similarity analysis for a few selected phonetic categories under both models we analyze in this study. 
Concretely, we retrieve five phonetic categories that exhibit the lowest JSD scores (and by implication the highest similarity) for each of the categories in the set \textipa{/w, E, S, g/}.
We then provide a ranking in the table from the most similar to the least. 
In the case of the approximant or semivowel  \textipa{/w/}, we observe that the approximant sound \textipa{/l/} exhibits the highest similarity under both models, but four vowels appear in ranks $2-5$. 
This indicates a high similarity in phonetic distributions between the approximant  \textipa{/w/} and vowels, which we further study in the clustering analysis below. 
For the front vowel \textipa{/E/}, the top-5 similar categories are all vowels under both models, although no strong preference for other front vowels can be observed since similar vowels are a mixture of front and central vowels. 
The two models exhibit the highest agreement in the case of the unvoiced post-alveolar fricative \textipa{/S/}, since both models have identical ranks that include the voiced post-alveolar fricative \textipa{/Z/} and the affricates \textipa{/tS, dZ/} among the most similar. 
For the voiced velar plosive \textipa{/g/}, the unvoiced velar plosive \textipa{/k/} is the most similar, as expected.

\subsection{Hierarchical clustering} 
To study the similarity patterns among the phonetic categories, we apply agglomerative hierarchical clustering with the Ward algorithm \cite{ward1963hierarchical} over the distance matrix generated by category-wise JSD values. 
The result of this clustering is illustrated in Fig.~\ref{fig:clustering}, where each phonetic category is colored by the manner of articulation. 
We observe that the clustering analysis yields a similar high-level grouping between \textsc{w2v2} and \textsc{xlsr}, except for the placement of nasals which differs across the two models.
For \textsc{w2v2} in Fig. \ref{fig:clustering}A, the highest level of organization divides the phonetic categories into two groups: a group that represents obstruent sounds (i.e., plosives, fricatives, and affricates) as well as nasals, and another group that represents vowels and approximants. 
On the other hand, the highest level of organization in \textsc{xlsr} in Fig. \ref{fig:clustering}B  reveals a pure obstruent vs. sonorant division, since both approximants and nasals exhibit a higher similarity to vowels than to other consonants.  
The consistent grouping of approximant sounds with vowels is not surprising given their acoustic-phonetic properties. 
Even though approximant sounds are considered consonants from a phonological point of view, they are produced with a (relatively) unconstricted articulation and exhibit a formant structure similar to vowels \cite{raphael2021acoustic}. 
 
Considering lower-level grouping  for obstruent consonants, labio-dental and dental fricatives \textipa{/f, v, T, D/} exhibit a higher similarity to plosive sounds \textipa{/p, b, t, d, k, g/} than alveolar and postalveolar fricatives  \textipa{/s, z, S, Z/} in both models. 
The affricates \textipa{/\textdyoghlig, \textteshlig/} are grouped together with alveolar and postalveolar fricatives under both models, indicating the prominence of the fricative component of affricates in their underlying distributions over the discrete units. 
The only phonetic category that exhibits unexpected behavior in this analysis is the glottal fricative \textipa{/h/}, which is grouped within plosives under both models.
However, the placement of the fricative \textipa{/h/} among plosives should not be surprising given that the voiceless plosives \textipa{/p, t, k/} are typically aspirated in syllable-initial position before a stressed vowel.
Plosive aspiration is acoustically realized as a friction noise following the release of the plosive, similar to the friction of the sound \textipa{/h/}.
Furthermore, the lowest level of grouping reflects the high similarity of phonetic minimal pairs (i.e., voicing contrasts)  among all plosive contrasts (i.e., \textipa{/t, d/}, \textipa{/p, b/}, and \textipa{/k, g/}), but only two fricative contrasts (i.e., \textipa{/s, z/} and  \textipa{/S, Z/}). 
As for the vowels, the lower-level grouping seems to reflect vowel backness more than vowel height in both models, although only a slight tendency to separate front vowels from back vowels can be observed.

\subsection{Correlation with feature-based phonetic distance} 
To study the degree to which our measure of (dis)similarity (JSD) reflects phonetic distance, we correlate the distance among phonetic distributions over discrete units against a measure of feature-based phonetic distance. 
To this end, we map each phonetic category in the TIMIT inventory onto a discrete, multi-valued feature vector based on the PHOIBLE  feature set \cite{moran2014phoible}. 
We then compute the feature-based distance as the Hamming distance between their feature vectors. 
When we consider all phonetic categories, we find a strong positive correlation between the JSD and feature-based phonetic distance in \textsc{w2v2} ($r = 0.63$) and \textsc{xlsr} ($r = 0.61$). 
Surprisingly, the correlation becomes stronger when we consider only the vowels in our analysis for both  \textsc{w2v2} ($r = 0.77$) and \textsc{xlsr} ($r = 0.80$), while it becomes weaker for consonants in \textsc{w2v2} ($r = 0.47$) and \textsc{xlsr} ($r = 0.43$). 
The weaker correlation among the consonants could be attributed to the high similarity between the phonetic distributions of vowels and approximants in both \textsc{w2v2} and \textsc{xlsr}, and vowels and nasals in \textsc{xlsr}. 
The correlation coefficients reported in this section are all Pearson's $r$ and significant with $p \ll 0.001$.

\section{Discussion and Conclusion}
We presented an information-theoretic framework for characterizing the relationship between phonetic categories and discrete units in self-supervised speech models. 
By representing each phonetic category as a distribution over discrete units, we have shown that the distribution entropy reflects the acoustic-phonetic variability of the underlying speech sounds, with vowels being more entropic on average than consonants. 
Moreover, phonetically similar sounds have been found to exhibit similar distributions, with the highest level of division separating obstruents and sonorants.
Our findings confirm the characterization of discrete units as sub-phonemic events, rather than high-level categories such as phonemes, which is consistent with the findings of Wells et al. \cite{wells2022phonetic}. 
Given that speech sounds are dynamic acoustic signals that vary considerably due to many factors such as context and speaker, we argue that the characterization of phonetic categories as distributions over sub-phonemic events allows for a more nuanced understanding of the relationships between phonetic categories and discrete units in self-supervised speech models.
Our presented analysis has a few limitations. 
For example, since we do not control for the different sources of variability of speech, it is difficult to disentangle the effect of these sources on the entropy of the phonetic distributions. 
Future work can further tackle this limitation with a controlled analysis with respect to the speaker and context variations. 

\section{Acknowledgements}
We thank the anonymous reviewers for their positive feedback. We extend our thanks to Marius Mosbach, Miaoran Zhang, and Vagrant Gautam for their comments on the paper.
This research is funded by the Deutsche Forschungsgemeinschaft (DFG, German Research Foundation), Project-ID 232722074 -- SFB 1102.

\bibliographystyle{IEEEtran}
\bibliography{paper}

\begin{thebibliography}{10}
\providecommand{\url}[1]{#1}
\csname url@samestyle\endcsname
\providecommand{\newblock}{\relax}
\providecommand{\bibinfo}[2]{#2}
\providecommand{\BIBentrySTDinterwordspacing}{\spaceskip=0pt\relax}
\providecommand{\BIBentryALTinterwordstretchfactor}{4}
\providecommand{\BIBentryALTinterwordspacing}{\spaceskip=\fontdimen2\font plus
\BIBentryALTinterwordstretchfactor\fontdimen3\font minus
  \fontdimen4\font\relax}
\providecommand{\BIBforeignlanguage}[2]{{%
\expandafter\ifx\csname l@#1\endcsname\relax
\typeout{** WARNING: IEEEtran.bst: No hyphenation pattern has been}%
\typeout{** loaded for the language `#1'. Using the pattern for}%
\typeout{** the default language instead.}%
\else
\language=\csname l@#1\endcsname
\fi
#2}}
\providecommand{\BIBdecl}{\relax}
\BIBdecl

\bibitem{9893562}
A.~Mohamed, H.-y. Lee, L.~Borgholt, J.~D. Havtorn, J.~Edin, C.~Igel,
  K.~Kirchhoff, S.-W. Li, K.~Livescu, L.~Maaløe, T.~N. Sainath, and
  S.~Watanabe, ``Self-supervised speech representation learning: A review,''
  \emph{IEEE Journal of Selected Topics in Signal Processing}, vol.~16, no.~6,
  pp. 1179--1210, 2022.

\bibitem{van2018representation}
A.~van~den Oord, Y.~Li, and O.~Vinyals, ``Representation learning with
  contrastive predictive coding,'' \emph{arXiv preprint arXiv:1807.03748},
  2018.

\bibitem{Schneider2019wav2vecUP}
S.~Schneider, A.~Baevski, R.~Collobert, and M.~Auli, ``wav2vec: Unsupervised
  pre-training for speech recognition,'' in \emph{Interspeech}, 2019.

\bibitem{baevski2020wav2vec}
A.~Baevski, Y.~Zhou, A.~Mohamed, and M.~Auli, ``wav2vec 2.0: A framework for
  self-supervised learning of speech representations,'' \emph{Advances in
  neural information processing systems}, vol.~33, pp. 12\,449--12\,460, 2020.

\bibitem{hsu2021hubert}
W.-N. Hsu, B.~Bolte, Y.-H.~H. Tsai, K.~Lakhotia, R.~Salakhutdinov, and
  A.~Mohamed, ``Hubert: Self-supervised speech representation learning by
  masked prediction of hidden units,'' \emph{IEEE/ACM Transactions on Audio,
  Speech, and Language Processing}, vol.~29, pp. 3451--3460, 2021.

\bibitem{devlin-etal-2019-bert}
\BIBentryALTinterwordspacing
J.~Devlin, M.-W. Chang, K.~Lee, and K.~Toutanova, ``{BERT}: Pre-training of
  deep bidirectional transformers for language understanding,'' in
  \emph{Proceedings of the 2019 Conference of the North {A}merican Chapter of
  the Association for Computational Linguistics: Human Language Technologies,
  Volume 1 (Long and Short Papers)}.\hskip 1em plus 0.5em minus 0.4em\relax
  Minneapolis, Minnesota: Association for Computational Linguistics, Jun. 2019,
  pp. 4171--4186. [Online]. Available: \url{https://aclanthology.org/N19-1423}
\BIBentrySTDinterwordspacing

\bibitem{higy2021discrete}
B.~Higy, L.~Gelderloos, A.~Alishahi, and G.~Chrupa{\l}a, ``Discrete
  representations in neural models of spoken language,'' in \emph{Proceedings
  of the Fourth BlackboxNLP Workshop on Analyzing and Interpreting Neural
  Networks for NLP}, 2021, pp. 163--176.

\bibitem{nguyen2022discrete}
T.~A. Nguyen, B.~Sagot, and E.~Dupoux, ``Are discrete units necessary for
  spoken language modeling?'' \emph{IEEE Journal of Selected Topics in Signal
  Processing}, vol.~16, no.~6, pp. 1415--1423, 2022.

\bibitem{wells2022phonetic}
D.~Wells, H.~Tang, and K.~Richmond, ``Phonetic analysis of self-supervised
  representations of english speech,'' in \emph{23rd Annual Conference of the
  International Speech Communication Association, INTERSPEECH 2022}.\hskip 1em
  plus 0.5em minus 0.4em\relax ISCA, 2022, pp. 3583--3587.

\bibitem{sicherman2023analysing}
A.~Sicherman and Y.~Adi, ``Analysing discrete self supervised speech
  representation for spoken language modeling,'' \emph{arXiv preprint
  arXiv:2301.00591}, 2023.

\bibitem{shannon2001mathematical}
C.~E. Shannon, ``A mathematical theory of communication,'' \emph{ACM SIGMOBILE
  mobile computing and communications review}, vol.~5, no.~1, pp. 3--55, 2001.

\bibitem{10.1162/tacl_a_00296}
\BIBentryALTinterwordspacing
T.~Pimentel, B.~Roark, and R.~Cotterell, ``{Phonotactic Complexity and Its
  Trade-offs},'' \emph{Transactions of the Association for Computational
  Linguistics}, vol.~8, pp. 1--18, 01 2020. [Online]. Available:
  \url{https://doi.org/10.1162/tacl\_a\_00296}
\BIBentrySTDinterwordspacing

\bibitem{pimentel-etal-2021-surprisal}
\BIBentryALTinterwordspacing
T.~Pimentel, C.~Meister, E.~Salesky, S.~Teufel, D.~Blasi, and R.~Cotterell, ``A
  surprisal{--}duration trade-off across and within the world{'}s languages,''
  in \emph{Proceedings of the 2021 Conference on Empirical Methods in Natural
  Language Processing}.\hskip 1em plus 0.5em minus 0.4em\relax Online and Punta
  Cana, Dominican Republic: Association for Computational Linguistics, Nov.
  2021, pp. 949--962. [Online]. Available:
  \url{https://aclanthology.org/2021.emnlp-main.73}
\BIBentrySTDinterwordspacing

\bibitem{rathi-etal-2021-information}
\BIBentryALTinterwordspacing
N.~Rathi, M.~Hahn, and R.~Futrell, ``An information-theoretic characterization
  of morphological fusion,'' in \emph{Proceedings of the 2021 Conference on
  Empirical Methods in Natural Language Processing}.\hskip 1em plus 0.5em minus
  0.4em\relax Online and Punta Cana, Dominican Republic: Association for
  Computational Linguistics, Nov. 2021, pp. 10\,115--10\,120. [Online].
  Available: \url{https://aclanthology.org/2021.emnlp-main.793}
\BIBentrySTDinterwordspacing

\bibitem{wu-etal-2019-morphological}
\BIBentryALTinterwordspacing
S.~Wu, R.~Cotterell, and T.~O{'}Donnell, ``Morphological irregularity
  correlates with frequency,'' in \emph{Proceedings of the 57th Annual Meeting
  of the Association for Computational Linguistics}.\hskip 1em plus 0.5em minus
  0.4em\relax Florence, Italy: Association for Computational Linguistics, Jul.
  2019, pp. 5117--5126. [Online]. Available:
  \url{https://aclanthology.org/P19-1505}
\BIBentrySTDinterwordspacing

\bibitem{Hahn2018AnIE}
M.~Hahn, J.~Degen, N.~D. Goodman, D.~Jurafsky, and R.~Futrell, ``An
  information-theoretic explanation of adjective ordering preferences,''
  \emph{Cognitive Science}, 2018.

\bibitem{futrell-etal-2015-quantifying}
\BIBentryALTinterwordspacing
R.~Futrell, K.~Mahowald, and E.~Gibson, ``Quantifying word order freedom in
  dependency corpora,'' in \emph{Proceedings of the Third International
  Conference on Dependency Linguistics (Depling 2015)}.\hskip 1em plus 0.5em
  minus 0.4em\relax Uppsala, Sweden: Uppsala University, Uppsala, Sweden, Aug.
  2015, pp. 91--100. [Online]. Available:
  \url{https://aclanthology.org/W15-2112}
\BIBentrySTDinterwordspacing

\bibitem{garofolo1993timit}
J.~S. Garofolo, ``{TIMIT} acoustic phonetic continuous speech corpus,''
  \emph{Linguistic Data Consortium, 1993}, 1993.

\bibitem{rasanen2016analyzing}
O.~R{\"a}s{\"a}nen, T.~Nagamine, and N.~Mesgarani, ``Analyzing distributional
  learning of phonemic categories in unsupervised deep neural networks,'' in
  \emph{CogSci... Annual Conference of the Cognitive Science Society. Cognitive
  Science Society (US). Conference}, vol. 2016.\hskip 1em plus 0.5em minus
  0.4em\relax NIH Public Access, 2016, p. 1757.

\bibitem{conneau2020unsupervised}
A.~Conneau, A.~Baevski, R.~Collobert, A.~Mohamed, and M.~Auli, ``Unsupervised
  cross-lingual representation learning for speech recognition,'' \emph{arXiv
  preprint arXiv:2006.13979}, 2020.

\bibitem{peterson1952control}
G.~E. Peterson and H.~L. Barney, ``Control methods used in a study of the
  vowels,'' \emph{The Journal of the acoustical society of America}, vol.~24,
  no.~2, pp. 175--184, 1952.

\bibitem{hillenbrand1995acoustic}
J.~Hillenbrand, L.~A. Getty, M.~J. Clark, and K.~Wheeler, ``Acoustic
  characteristics of american english vowels,'' \emph{The Journal of the
  Acoustical society of America}, vol.~97, no.~5, pp. 3099--3111, 1995.

\bibitem{hagiwara1997dialect}
R.~Hagiwara, ``Dialect variation and formant frequency: The american english
  vowels revisited,'' \emph{The Journal of the Acoustical Society of America},
  vol. 102, no.~1, pp. 655--658, 1997.

\bibitem{ward1963hierarchical}
J.~H. Ward~Jr, ``Hierarchical grouping to optimize an objective function,''
  \emph{Journal of the American statistical association}, vol.~58, no. 301, pp.
  236--244, 1963.

\bibitem{raphael2021acoustic}
L.~J. Raphael, ``Acoustic cues to the perception of segmental phonemes,''
  \emph{The handbook of speech perception}, pp. 603--631, 2021.

\bibitem{moran2014phoible}
\BIBentryALTinterwordspacing
S.~Moran and D.~McCloy, Eds., \emph{PHOIBLE 2.0}.\hskip 1em plus 0.5em minus
  0.4em\relax Jena: Max Planck Institute for the Science of Human History,
  2019. [Online]. Available: \url{https://phoible.org/}
\BIBentrySTDinterwordspacing

\end{thebibliography}

\end{document}